\documentclass[runningheads]{llncs}
\usepackage[utf8]{inputenc}
\usepackage[british]{babel}
\usepackage{csquotes}
\usepackage{graphicx}
\usepackage[caption=false]{subfig}
\usepackage{array}
\usepackage{stfloats}

\usepackage[hidelinks]{hyperref}
\usepackage[misc,geometry]{ifsym}

\usepackage{color}

\newcommand{\iam}[0]{$IAM_{synth}$}
\newcommand{\draculaReal}[0]{$Dracula_{real}$}
\newcommand{\draculaSynth}[0]{$Dracula_{synth}$}

\makeatletter
\AtBeginDocument{%
  \def\doi#1{\url{https://doi.org/#1}}}
\makeatother

\begin{document}
\title{Paired Image to Image Translation for Strikethrough Removal From Handwritten Words}
\titlerunning{Paired Image Strikethrough Removal}
%
%
\author{Raphaela Heil\inst{1}(\Letter)\orcidID{0000-0002-5010-9149} \and Ekta Vats\inst{2}\orcidID{0000-0003-4480-3158} \and Anders Hast\inst{1}\orcidID{0000-0003-1054-2754}}
\authorrunning{R. Heil, et al.}
%
\institute{Division for Visual Information and Interaction, \\Department of Information Technology,\\ Uppsala University, Uppsala, Sweden\\
\email{\{raphaela.heil, anders.hast\}@it.uu.se}\\
\and Centre for Digital Humanities Uppsala,\\Department of Archives, Libraries and Museums,\\ Uppsala Universtiy, Uppsala, Sweden \\ \email{ekta.vats@abm.uu.se}}
\maketitle              
\begin{abstract}
Transcribing struck-through, handwritten words, for example for the purpose of genetic criticism, can pose a challenge to both humans and machines, due to the obstructive properties of the superimposed strokes. This paper investigates the use of paired image to image translation approaches to remove strikethrough strokes from handwritten words. Four different neural network architectures are examined, ranging from a few simple convolutional layers to deeper ones, employing Dense blocks. Experimental results, obtained from one synthetic and one genuine paired strikethrough dataset, confirm that the proposed paired models outperform the CycleGAN-based state of the art, while using less than a sixth of the trainable parameters.

\keywords{Strikethrough Removal  \and Paired Image to Image Translation \and Handwritten Words \and Document Image Processing.}
\end{abstract}

\section{Introduction}
Struck-through words generally appear at a comparably low frequency in different kinds of handwritten documents. Regardless of this, reading what was once written and subsequently struck through can be of interest to scholars from the humanities, such as literature, history and genealogy \cite{van_hulle}. In order to facilitate strikethrough-related research questions from such fields, strikethrough removal approaches have been proposed in the area of document image analysis \cite{CHAUDHURI2017282,we,texrgan}. 

In this work, we approach the problem from a new perspective, employing a paired image to image translation method. The underlying idea of this family of models consists of the use of two corresponding images, one from the source and one from the target domain \cite{pix2pix}. Using these paired images, deep neural networks are trained to learn the transformation from the source to the target domain. As indicated by the name, this approach is generally limited by the availability of paired data. This poses a particular challenge when applying paired approaches to the task of strikethrough removal. Once a word has been struck through, the original, clean word is permanently altered and effectively becomes unobtainable from the manuscript itself. In order to mitigate this problem, we examine the paired image to image translation setting using a combination of synthetic and genuine data. The general approach is implemented and examined via a selection of four deep neural networks of varying size and architectural complexity. All models are evaluated using a synthetic and a genuine test dataset \cite{we}.


The main contributions of this work are as follows. This work advances the state of the art in strikethrough removal by using a paired image to image translation approach. To the best of the authors' knowledge, this is the first attempt at using such an approach towards strikethrough removal. In order to overcome the issue of insufficient paired data, this work uses a combination of synthetic and genuine datasets, and also introduces a new dataset (\draculaSynth) to further supplement the training data. Four different models are evaluated and compared with the CycleGAN-based state of the art \cite{we}, using two existing datasets \cite{iam_strikethrough,dracula_strikethrough} and the new \draculaSynth \ dataset. The \draculaSynth \ dataset, the source code and the pre-trained models will be made publicly available.




\begin{figure}[!t]
    \centering
    \subfloat[]{\includegraphics[height=1.5cm]{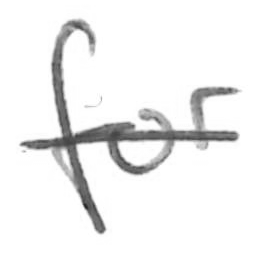}%
}
\hfil
\subfloat[]{\includegraphics[height=1.5cm]{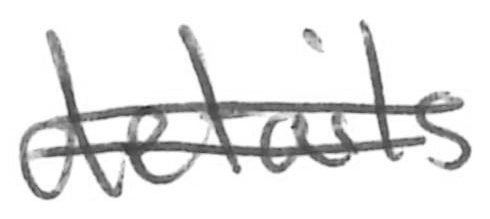}%
}
\hfil
\subfloat[]{\includegraphics[height=1.5cm]{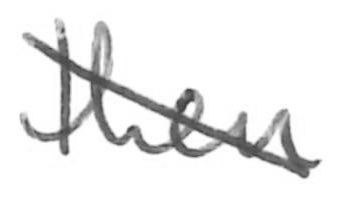}%
}
\hfil
\subfloat[]{\includegraphics[height=1.5cm]{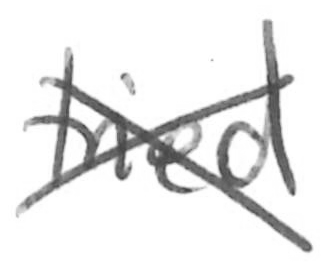}%
}
\hfil\\
\subfloat[]{\includegraphics[height=1.5cm]{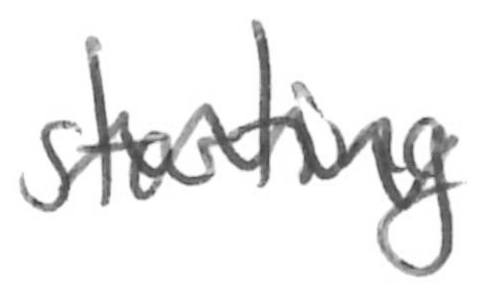}%
}
\hfil
\subfloat[]{\includegraphics[height=1.5cm]{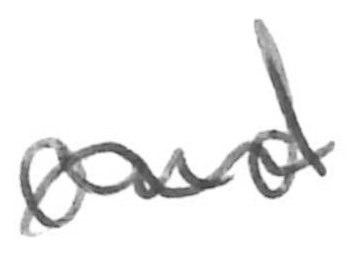}%
}
\hfil
\subfloat[]{\includegraphics[height=1.5cm]{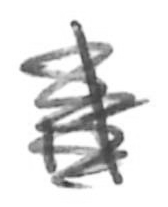}%
}
\caption{Examples for the types of strikethrough strokes considered in this work. a) Single, b) Double, c) Diagonal, d) Cross, e) Zig Zag, f) Wave, g) Scratch. All samples are taken from the test set of \draculaReal, i.e. displaying genuine strikethrough stokes.}
\label{fig:samples}
\end{figure}

\section{Related Works}

\subsection{Strikethrough Processing}
At the time of writing, three approaches for removing strikethrough strokes from handwritten words have been proposed. Firstly, Chaudhuri et al. introduce a graph-based approach, identifying edges belonging to the strikethrough stroke and removing them, using inpainting \cite{CHAUDHURI2017282}. They report $F_1$ scores of 0.9116 on a custom, unbalanced database, containing strikethrough strokes of the types \textit{single}, \textit{multiple}, \textit{slanted}, \textit{crossed}, \textit{zig zag} and \textit{wavy} (cf. \autoref{fig:samples}). Chaudhuri et al. do not consider the stroke type that we denote \textit{scratch}.

In addition to this, Poddar et al. \cite{texrgan} employ a semi-supervised, generative adversarial network (GAN), which is trained using a combination of GAN-loss, Structural Similarity Index and $L_1$-norm. They report an average $F_1$ score of 0.9676 on their own synthetic IAM database, which contains strokes of types \textit{straight}, \textit{slanted}, \textit{crossed}, \textit{multiple straight} and \textit{partial straight} and \textit{partial slanted} strokes. In contrast to Poddar et al., we also consider strikethrough strokes of the types \textit{wave}, \textit{zig zag} and \textit{scratch} in this work. As can be seen in \autoref{fig:samples}, the latter strokes are considerably more challenging that the former ones. 

Lastly, Heil et al. investigate the use of attribute-guided cycle consistent GANs (CycleGANs) \cite{we}. They report $F_1$ scores of up to 0.8172, respectively 0.7376, for their synthetic and genuine test sets. In contrast to the two former approaches, the datasets and code from Heil et al. are publicly available \cite{code_cycle,iam_strikethrough,dracula_strikethrough}. We reuse their datasets in this work as the basis for our experiments as well as to compare our results to that of \cite{we}. \\

Besides the removal of strikethrough, a number of works have concerned themselves with other aspects related to the processing of struck-through words. Various approaches \cite{CHAUDHURI2017282,we,detection,sukalpa}, have been proposed to classify whether a given image depicts a struck-through or clean word. In addition to this, a number of works have examined the extent to which strikethrough impacts the performance of writer identification \cite{adak_writer} and text and character recognition approaches \cite{schomaker,likforman_hmm,nisa_htr}. Some of the aforementioned works are briefly summarised in a survey from early 2021 by Dheemanth Urs and Chethan \cite{str_survey}.

\subsection{Paired Image to Image Translation}
A variety of approaches for paired image to image translation have been proposed over the years. One prominent example, employing a conditional GAN, is \textit{Pix2Pix} \cite{pix2pix}. In the field of document image analysis, paired image to image translation approaches have for example been used in the form of auto-encoders to remove various types of noise \cite{adversarial_autoencoder,officeAutoencoder}, as well as to binarise manuscript images \cite{CALVOZARAGOZA201937,palm}. To the best of our knowledge, paired image to image translation approaches have not yet been used for the task of strikethrough removal.

\subsection{Strikethrough Datasets}
While strikethrough occurs naturally in a variety of real-world datasets, such as the IAM database \cite{iam}, the amount and diversity are generally too small to be used to efficiently train larger neural networks. At the time of writing, only two datasets, which focus specifically on strikethrough-related research questions, are publicly available. The first one, which we will refer to as \iam \cite{iam_strikethrough} in this work, is an IAM-based \cite{iam} dataset consisting of genuine handwritten words that have been altered with synthetic strikethrough strokes \cite{we}. In addition to this, a smaller, but genuine, dataset of struck-through, handwritten words exists. This single-writer dataset, which we will refer to as \draculaReal \cite{dracula_strikethrough}, contains handwritten word images and their struck-through counterparts from Bram Stoker's \textit{Dracula}. Heil et al. \cite{we} collected the genuine dataset by scanning and aligning handwritten words, via \cite{johan}, before and after the strikethrough was applied.

\section{Image to Image Translation Models for Strikethrough Removal}
In contrast to earlier works \cite{we,texrgan}, we approach the task of strikethrough removal using paired image to image translation. For this, a model receives an image from the source domain as input and is trained to reproduce the image as it would appear in the target domain. In this work, we propose to consider struck-through images as the source and cleaned images, i.e. without strikethrough strokes, as the target domain.


We examine four different deep neural network architectures in the paired image to image translation setting and compare their performance to the attribute-guided CycleGANs proposed by Heil et al. \cite{we}. The chosen architectures were selected with the aim of exploring models with a range of layer arrangements and varying amounts of trainable parameters. A brief overview of the architectures, and the names by which we will refer to them for the remainder of this work, are given below. Furthermore, \autoref{fig:model_vis} presents a schematic summary. For a more detailed description of the architectures, such as number of convolutional filters per layer, stride, padding, etc., the interested reader is referred to the code, accompanying this paper (cf. \autoref{sec:code} - Dataset and Code Availability).

\paragraph{SimpleCNN} As indicated by the name, this model constitutes a simple  convolutional neural network, consisting of three up- and down-sampling layers with few (16, respectively 32) filters.

\paragraph{Shallow} This network consists of two convolutional down-, respectively up-sampling layers. It does not contain any intermediate bottleneck layers. The arrangement of layers in this architecture is the same as the outer layers in the \textit{Generator} one.

\paragraph{UNet} This dense UNet \cite{tiramisu} consists of one down- and up-sampling, as well as one bottleneck block, each of which consisting of a Dense block with four dense layers. The two outer dense blocks are furthermore connected via a skip connection. We use the implementation provided by \cite{octopytorch}.

\paragraph{Generator} This network uses the same architecture as the generator that was used in the attribute-guided CycleGAN by Heil et al. \cite{we}. It consists of three convolutional up-, respectively down-sampling layers, with a Dense \cite{densenet} bottleneck in between. Each of the convolutional layers is followed by a batch normalisation step, as well as a rectified linear unit (ReLU). \\

With the exception of \textit{UNet}, all of the models above use a sigmoid as the final activation function. For \textit{UNet}, we follow the original implementation, provided by \textit{OctoPyTorch} \cite{octopytorch}, which uses the identity function as activation and combines the network with a \textit{binary cross entropy with logits loss}. 

\autoref{tab:exp_models} summarises the trainable parameter counts for the described models, as well as the attribute-guided CycleGAN from \cite{we}. As can be seen from the table the chosen models cover a considerable range of trainable parameter amounts. 

\begin{figure}[!b]
    \centering
    \includegraphics[width=\textwidth]{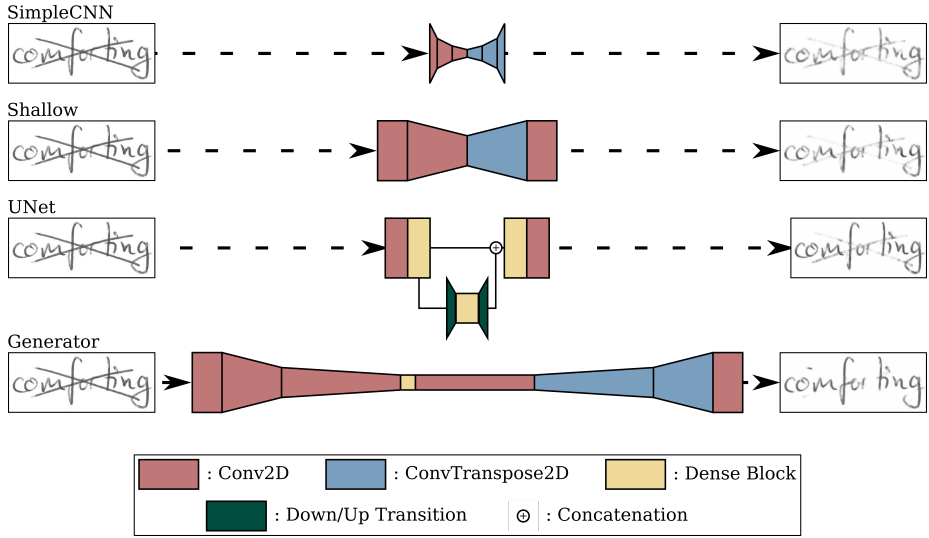}
    \caption{Schematic overview over the four paired architectures examined in this work. Red boxes are regular 2D convolutional layers, blue transposed convolutions, green up-, respectively down-sampling layers and yellow ones dense blocks. Height of the boxes represents an approximate measure of the input, respectively output size, while the width indicates the number of convolutional filters (Conv2D and ConvTranspose2D), respectively number of dense layers (Dense Block).}
    \label{fig:model_vis}
\end{figure}

\setlength{\tabcolsep}{0.3em}
\begin{table}[!b]
    \centering
    \caption{Number of trainable parameters per model. It should be noted that the CycleGAN parameter count includes the generators \textit{and} discriminators but not the pre-trained auxiliary discriminator.}
    \begin{tabular}{|c|c|}
        \hline
        Model Name & Parameter Count \\
         \hline
         \hline
        SimpleCNN & 28\,065\\
        \hline
        Shallow & 154\,241\\
        \hline
        UNet & 181\,585\\
        \hline 
        Generator & 1\,345\,217\\
        \hline
        \hline
        Attribute-guided CycleGAN \cite{we} & 8\,217\,604 \\
        \hline
    \end{tabular}
    \label{tab:exp_models}
\end{table}

\section{Experiment Setup}

\subsection{Datasets}
As mentioned above, at the time of writing, two strikethrough-related datasets, \iam \ and \draculaReal, are publicly available. We base our experiments on these and introduce an additional synthetic dataset. The latter one, which we will refer to as \draculaSynth, consists of the clean images from the \draculaReal \ training split, to which synthetic strikethrough was applied via the method proposed in \cite{we}. We repeat the generation process five times, using different seed values for the random number generator, resulting in five different strikethrough strokes for each individual word image. In addition to combining all of these generated images into one large training set, we also consider the separate partitions, numbered 0 to 4, each of which contains only one instance of any given word image. It should be noted that \draculaSynth \ is used exclusively during the training phase. For validation and testing, the original images from \draculaReal \ are used. However, if necessary for future investigations, synthetic strikethrough can also be generated for the original validation and test words, using the approach outlined above. \autoref{tab:datasets} summarises the three described datasets with respect to the number of images per split and the number of writers. 

\setlength{\tabcolsep}{0.3em}
\begin{table}[!t]
    \centering
    \caption{Summary of the three datasets used in this work. Numbers indicate the amount of words contained in each dataset split.}
    \begin{tabular}{|c|c|c|c|c|}
    \hline
        Dataset & Train & Validation & Test & Multi-writer\\
        \hline 
        \hline
         \iam & 3066 & 273 & 819 & yes\\
         \hline
        \draculaReal  & 126 & 126 & 378 & no\\
        \hline
        \draculaSynth  & 5 x 126 & N/A & N/A & no\\
        \hline
    \end{tabular}
    \label{tab:datasets}
\end{table}

\subsection{Neural Network Training Protocol}
All models are implemented in PyTorch 1.7 \cite{pytorch} and are trained for a total of 30 epochs with a batch size of four, using the Adam \cite{adam} optimiser with default parameters. We use a regular binary cross entropy loss for all models, except \textit{UNet}, for which the binary cross entropy with logits loss is used, as mentioned above. Each model is trained from scratch using each of the following datasets: 

\begin{itemize}
    \item \iam \ training split
    \item \draculaSynth \ training partitions 0-4, each individually
    \item a combination of all \draculaSynth \ training partitions, i.e. 630 images
\end{itemize}

For all of the models, we monitor the performance, measured via the $F_1$ score, using the validation split of the respective dataset (\iam \ or \draculaReal). We retain each model's weights from the best performing epoch for the evaluation on the respective test split. Each model is separately retrained 30 times, yielding 30 sets of weights per model and dataset combination. 

In addition to the models described above, we also train a number of CycleGANs, following the procedure described in \cite{we}. As above, each model is retrained 30 times. 

Regardless of architecture, all images are, where necessary, converted to greyscale, inverted so that ink pixels have the highest intensities in the image, and are scaled to a height of 128 and padded with the background colour (black) to a fixed width of 512.

\subsection{Evaluation Protocol}
Each of the trained models is applied to the struck-through images from the test split of \draculaReal. The resulting image is inverted and rescaled to its original dimensions and additional padding is removed, where applicable. Subsequently, the cleaned image is compared with the corresponding ground-truth, calculating the \textit{Root Mean Square Error} (RMSE) of the greyscale images, as well as the $F_1$ score of the Otsu-binarised \cite{otsu} ones. We calculate the $F_1$ score following the formula provided by \cite{CHAUDHURI2017282}, which is repeated below:

\begin{equation}
    Detection Rate (DR) = \frac{O2O}{N}
\end{equation}

\begin{equation}
    Recognition Accuracy (RA) = \frac{O2O}{M}
\end{equation}

\begin{equation}
    F_1 = \frac{2*DR*RA}{DR+RA}
\end{equation}

\noindent where $M$ is the number of ink pixels in the image, cleaned by the respective model; $N$ the number of ink pixels in the ground-truth image, and $O2O$ the number of matching pixels between the two images. 

Unless otherwise stated, we report the average and standard deviation, summarised over 30 repeated training runs for each of the model and dataset combinations.

\section{Results and Analysis}

In the following sections, the performances of the four paired image to image translation models are presented, compared with the \textit{CycleGAN}'s results and analysed. Besides the quantitative evaluation in subsections \ref{sec:iam} to \ref{sec:dracula_all}, a number of qualitative results are also presented in \autoref{sec:qualitative}.

\subsection{Models trained on \iam}\label{sec:iam}

In a first step, we compare the four paired models with the unpaired CycleGAN approach, when trained on the \iam \ dataset and evaluated on the \iam \  and \draculaReal \ test splits. Tables \ref{tab:iam_iam} and \ref{tab:iam_dracula} present the respective mean $F_1$ and \textit{RMSEs} scores for the five architectures, summarising the 30 repeated training runs. Values reported for \cite{we} stem from our own experiments, using the code by Heil et al. \cite{code_cycle}. 

As can be seen from \autoref{tab:iam_iam}, there is some degree of variation between the four paired models, when evaluating them on the \iam \ test split. However, all of them outperform the attribute-guided CycleGAN by a considerable margin. For the $F_1$ score, improvements range from 7.4 percentage points (pp) for the \textit{SimpleCNN}, up to roughly 17 pp for the \textit{Generator}. Similarly, the former model outperforms the \textit{CycleGAN} by approximately 4 pp with respect to the \textit{RMSE}, while the latter achieves a performance improvement of 9 pp. Comparing the four paired approaches with each other, the two larger models \textit{Generator} and \textit{UNet} prominently outperform the two smaller ones. 

Considering the model performances on the \draculaReal \ test split, shown in \autoref{tab:iam_dracula}, it can be noted that the four paired architectures do not drastically differ from each other. For the $F_1$ score, the \textit{SimpleCNN} and \textit{Generator} constitute the lower and upper bounds, respectively. Interestingly, with regard to the \textit{RMSE}, \textit{SimpleCNN} performs slightly better than \textit{Generator}, which ranks second in the overall comparison of values. A possible explanation for this could be that \textit{Generator} leaves behind traces of strikethrough which are removed by the binarisation, required for the $F_1$ score, but remain visible and make an impact in the \textit{RMSE} calculation. 

In the overall ranking, the \textit{CycleGAN} reaches the fifth ($F_1$), respectively third position (\textit{RMSE}), consistently being outperformed by the \textit{Generator} model with a margin of 3.8 pp for the $F_1$ score and roughly 0.5 pp for the \textit{RMSE}. Although this gain in cleaning performance is relatively small, the \textit{Generator} includes the additional benefit of having less than a sixth of the \textit{CycleGAN}'s trainable parameters. Taking the amount of trainable parameters further into account, the \textit{Shallow} and \textit{UNet} models provide a good trade-off between performance and size in this evaluation setting.

\begin{table}[!b]
    \centering
    \caption{Mean $F_1$ scores (higher better, range [0,1]) and \textit{RMSE}s (lower better, range [0,1]) for the five architectures, trained and evaluated on the train and test splits, respectively, of \iam. Standard deviation over thirty training runs given in parentheses. Best model marked in bold.}
    \begin{tabular}{|c|c|c|}
    \hline 
        Model & $F_1$ & RMSE\\
        \hline \hline
SimpleCNN & 0.8727 ($\pm$ 0.0042) & 0.0753 ($\pm$ 0.0025) \\
\hline
Shallow & 0.9163 ($\pm$ 0.0045) & 0.0558 ($\pm$ 0.0025) \\
\hline
UNet & 0.9599 ($\pm$ 0.0015) & 0.0301 ($\pm$ 0.0012) \\
\hline
Generator & \textbf{0.9697 ($\pm$ 0.0012)} & \textbf{0.0237 ($\pm$ 0.0016)} \\
\hline
\hline
Attribute-guided CycleGAN \cite{we} & 0.7981 ($\pm$ 0.0284) & 0.1172 ($\pm$ 0.0286) \\
\hline
    \end{tabular}
    \label{tab:iam_iam}
\end{table}

\begin{table}[!t]
    \centering
    \caption{Mean $F_1$ scores (higher better, range [0,1]) and \textit{RMSE}s (lower better, range [0,1]) for the five architectures, trained on \iam \ and evaluated on the test split of \draculaReal. Standard deviation over thirty training runs given in parentheses. Best model marked in bold.}
    \begin{tabular}{|c|c|c|}
    \hline 
        Model & $F_1$ & RMSE\\
        \hline \hline
SimpleCNN & 0.7204 ($\pm$ 0.0303) & \textbf{0.0827 ($\pm$ 0.0038)} \\
\hline
Shallow & 0.7450 ($\pm$ 0.0028) & 0.0932 ($\pm$ 0.0044) \\
\hline
UNet & 0.7451 ($\pm$ 0.0013) & 0.1005 ($\pm$ 0.0033) \\
\hline
Generator & \textbf{0.7577 ($\pm$ 0.0035)} & 0.0868 ($\pm$ 0.0021) \\
\hline
\hline
Attribute-guided CycleGAN \cite{we} & 0.7189 ($\pm$ 0.0243) & 0.0927 ($\pm$ 0.0212) \\
\hline
    \end{tabular}
    \label{tab:iam_dracula}
\end{table}

\begin{table}[!t]
    \centering
    \caption{Mean $F_1$ scores (higher better, range [0,1]) and \textit{RMSE}s (lower better, range [0,1]) for the five architectures, trained on individual partitions of \draculaSynth \ and evaluated on the test split of \draculaReal. Standard deviations over five partitions with thirty training runs each, given in parentheses. Best model marked in bold. }
    \begin{tabular}{|c|c|c|}
    \hline
    Model & $F_1$ & RMSE \\
    \hline \hline
SimpleCNN & 0.7327 ($\pm$ 0.0046) & 0.0757 ($\pm$ 0.0008) \\
\hline
Shallow & 0.7648 ($\pm$ 0.0052) & 0.0709 ($\pm$ 0.0023) \\
\hline
UNet & 0.7482 ($\pm$ 0.0031) & 0.0761 ($\pm$ 0.0047) \\
\hline
Generator & \textbf{0.7872 ($\pm$ 0.0059)} & \textbf{0.0655 ($\pm$ 0.0026)} \\
\hline
\hline
Attribute-guided CycleGAN \cite{we} & 0.5073 ($\pm$ 0.1484) & 0.1317 ($\pm$ 0.0312) \\
\hline

    \end{tabular}
    
    \label{tab:folds}
\end{table}

\begin{table}[!t]
    \centering
    \caption{Mean $F_1$ scores (higher better, range [0,1] and RMSEs (lower better, range [0,1]) for the five architectures, trained on the aggregated partitions of \draculaSynth \ and evaluated on the test split of \draculaReal. Standard deviation over thirty training runs given in parentheses. Best model marked in bold. }
    \begin{tabular}{|c|c|c|}
    \hline
    Model & $F_1$ & RMSE \\
    \hline \hline
SimpleCNN & 0.7543 ($\pm$ 0.0034) & 0.0718 ($\pm$ 0.0011) \\
\hline
Shallow & 0.7825 ($\pm$ 0.0049) & 0.0681 ($\pm$ 0.0029) \\
\hline
UNet & 0.7662 ($\pm$ 0.0064) & 0.0734 ($\pm$ 0.0038) \\
\hline
Generator & \textbf{0.8122 ($\pm$ 0.0031)} & \textbf{0.0592 ($\pm$ 0.0015)} \\
\hline
\hline
Attribute-guided CycleGAN \cite{we} & 0.6788 ($\pm$ 0.0516) & 0.1148 ($\pm$ 0.0373) \\
\hline

    \end{tabular}
    \label{tab:all}
\end{table}

\subsection{Models trained on Individual Partitions of \draculaSynth}

\autoref{tab:folds} summarises the performances for the five architectures trained on the five individual partitions of synthetic data from \draculaSynth, that were generated based on the train split of \draculaReal. As can be seen from the table, there is a considerable difference in performance between the paired models and the attribute-guided \textit{CycleGAN}. Again, the \textit{Generator} models perform best in comparison to the other paired approaches. It can also be noted that, with the exception of the \textit{CycleGAN}, on average all models trained on the \draculaSynth \ training partitions outperform their counterparts trained on \iam (cf. \autoref{tab:iam_dracula}).

\subsection{Models Trained on the Aggregation of Partitions from \draculaSynth}\label{sec:dracula_all}

Following the large improvements gained by training on the individual partitions of \draculaSynth, all models were retrained from scratch on the aggregation of the five partitions. This aggregating step was taken in order to investigate the impact of a more diverse set of strikethrough strokes, applied to handwriting from the target domain. \autoref{tab:all} shows the resulting $F_1$ and \textit{RMSE} scores for this experiment. The accumulated dataset yields moderate, yet consistent, increases for all models. Although a more substantial improvement of 17 pp for the $F_1$ score can be observed for the attribute-guided \textit{CycleGAN}, it still performs considerably worse than the paired approaches. Future experiments may investigate the impact of further increasing the size and diversity of synthetic datasets based on clean images from the target handwriting domain.



\begin{figure}[!b]
    \centering
    \includegraphics[height=6.8cm]{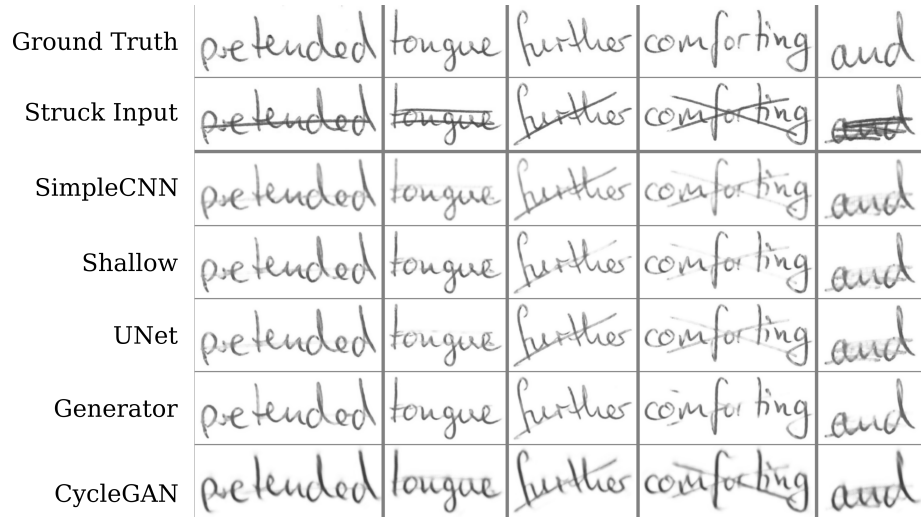}
    \caption{Cherry-picked examples for the five models. All images are taken from the \draculaReal \ test split and were processed by the respective model. Results are shown as the mean greyscale images, averaged over 30 model repetitions.}
    \label{fig:cherries}
\end{figure}

\subsection{Qualitative Results}\label{sec:qualitative}
\begin{figure}[!t]
    \centering
    \includegraphics[height=6.8cm]{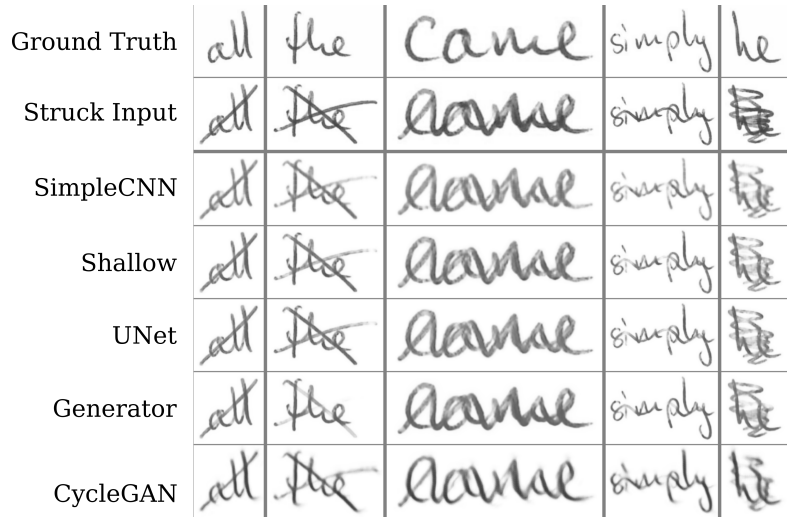} %
    \caption{Lemon-picked examples for the five models. All images are taken from the \draculaReal \ test split and were processed by the respective model. Results are shown as the mean greyscale images, averaged over 30 model repetitions.}
    \label{fig:lemons}
\end{figure}

In order to demonstrate the range of strikethrough removal capabilities of the evaluated models, we present a number of hand-picked positive (\enquote{cherry-picked}) and negative (\enquote{lemon-picked}) cases from the \draculaReal \ test split. These images are shown in \autoref{fig:cherries} and \autoref{fig:lemons}, respectively, and were obtained by calculating the mean of the 30 greyscale model outputs for each architecture, trained on the aggregated partitions of \draculaSynth. As can be seen from \autoref{fig:cherries}, most of the models manage to remove a fair portion of the genuine strikethrough, despite being only trained on synthetic strikethrough strokes. Additionally, it can be noted that the mean images for the paired approaches are generally more crisp than those obtained from \textit{CycleGAN}, which appear more blurry, indicating less agreement between individual model checkpoints. 

In contrast to the figure above, \autoref{fig:lemons} depicts mean images for lemon-picked examples. Some of the models remove portions of the strokes, for example, the majority of one of the diagonal strokes in the cross sample cleaned by the \textit{Generator} (second to last row, second image column, word \enquote{the}). 

Overall, inspecting the rest of the cleaned images, not pictured here for brevity, a general trend can be noted for different types of strikethrough strokes. Generally, most of the \textit{single}, \textit{double} and \textit{diagonal} strokes are removed convincingly. Shorter, \textit{scratched} out words are often cleaned less than their longer counterparts. Strokes of types \textit{cross}, \textit{zig zag} and \textit{wave} are cleaned considerably less often than the other stroke types.

\section{Conclusions}
In this work we have examined four paired image to image translation models and compared them with a state of the art unpaired strikethrough removal approach \cite{we}. Based on the presented results and analyses, we draw the following conclusions: 
\begin{enumerate}
    \item Paired image to image translation approaches outperform the attribute-guided CycleGAN, proposed by \cite{we}, in all of the experiments presented in this work. The examined models not only outperform the state of the art in terms of strikethrough removal performance but also contain considerably fewer trainable parameters, making them cheaper and faster to train per epoch. Although the best results are obtained from the largest paired approach, the smaller evaluated models still display a considerable cleaning performance and may therefore still be of interest in scenarios with limited computing resources.
    \item Using the custom synthetic strikethrough dataset, based on clean words from the same writer as the target domain (i.e. here \draculaSynth \ for \draculaReal), yields better results in a paired image to image translation setting than the much more diverse \iam \ dataset under the same experiment conditions. 
    \item Upon inspection of the qualitative results, stroke types can be separated into two groups, based on level of difficulty. \textit{Single}, \textit{double} and \textit{diagonal} lines are generally cleaned more easily than \textit{crosses}, \textit{zig zag} and \textit{waves}. For \textit{scratched} out words, a trend can be noted, indicating that longer words are cleaned more easily than shorter ones. 
\end{enumerate}

Overall, considering the use case of strikethrough removal in an archival context, for example as preprocessing step for genetic criticism, where one or few handwriting styles are present in the data, we recommend to explore options to create an in-domain, synthetic dataset, similar to the approach that was taken for \draculaSynth. In the presented case, slightly more than 100 clean words from the target handwriting style, combined with synthetic strikethrough, yielded models with considerable strikethrough removal abilities, generating convincingly cleaned words.

In the future, we aim to expand the scope of genuine strikethrough removal datasets in order to further investigate a variety of approaches and the impact of more diverse handwriting styles. \\

\noindent\textbf{Acknowledgements.} 
The computations were enabled by resources provided by the Swedish National Infrastructure for Computing (SNIC) at Chalmers Centre for Computational Science and Engineering (C3SE) partially funded by the Swedish Research Council through grant agreement no. 2018-05973. This work is partially supported by Riksbankens Jubileumsfond (RJ) (Dnr P19-0103:1).

\appendix
\section{Dataset and Code Availability}\label{sec:code}
\draculaSynth : \url{https://doi.org/10.5281/zenodo.6406538}\\
Code: \url{https://doi.org/10.5281/zenodo.6406284}.


%
%
%
\bibliographystyle{splncs04}
\bibliography{main}

\end{document}